\documentclass[letterpaper, 10 pt, conference]{ieeeconf}  

\IEEEoverridecommandlockouts                              

\usepackage{graphicx}
\usepackage{siunitx}
\usepackage{amsmath}
\usepackage{placeins}
\usepackage{booktabs}

\title{\LARGE \bf
A Passive Elastic-Folding Mechanism for Stackable Airdrop Sensors}

\author{Damyon Kim$^{1*}$, Yuichi Honjo$^{1}$, Tatsuya Iizuka$^{2}$, Naomi Okubo$^{1}$, Naoto Endo$^{2}$, \\ Hiroshi Matsubara$^{2}$, Yoshihiro Kawahara$^{1}$, Naoto Morita$^{1}$, and Takuya Sasatani$^{1*}$
\thanks{
This work was supported by NTT Space Environment and Energy Laboratories, JST FOREST Program (Grant Number JPMJFR242P), and JSPS KAKENHI (Grant Number 23K28068). Generative AI tools (OpenAI GPT and Google Gemini) were used to refine language and grammar.\newline\newline
$^{1}$Graduate School of Engineering, The University of Tokyo, Japan\newline
$^{2}$NTT Space Environment and Energy Laboratories, Japan\newline
$^{*}$Correspondence: sasatani@g.ecc.u-tokyo.ac.jp, kim\_d@akg.t.u-tokyo.ac.jp%
}}
\begin{document}

\maketitle
\thispagestyle{empty}
\pagestyle{empty}

\begin{abstract}
Air-dispersed sensor networks deployed from aerial robotic systems (\textit{e.g.}, UAVs) provide a low-cost approach to wide-area environmental monitoring.
However, existing methods often rely on active actuators for mid-air shape or trajectory control, increasing both power consumption and system cost.
Here, we introduce a passive elastic-folding hinge mechanism that transforms sensors from a flat, stackable form into a three-dimensional structure upon release.
Hinges are fabricated by laminating commercial sheet materials with rigid printed circuit boards (PCBs) and programming fold angles through a single oven-heating step, enabling scalable production without specialized equipment.
Our geometric model links laminate geometry, hinge mechanics, and resulting fold angle, providing a predictive design methodology for target configurations.
Laboratory tests confirmed fold angles between $10^\circ$ and $100^\circ$, with a standard deviation of $4^\circ$ and high repeatability.
Field trials further demonstrated reliable data collection and LoRa transmission during dispersion, while the Horizontal Wind Model (HWM)-based trajectory simulations indicated strong potential for wide-area sensing exceeding 10 km.

\end{abstract}

\section{INTRODUCTION}

Air-dispersed sensor systems deployed from aerial robotic platforms such as UAVs and High-Altitude Platform Stations (HAPS) enable wide-area environmental monitoring~\cite{iyer_airdropping_2020, iyer_wind_2022, wiesemuller_transient_2022}.
Aerial platforms deliver sensors to distant sites, while the low cost of distributed nodes supports dense, scalable coverage.
The descent trajectory of these sensors—and thus the sensing range—is strongly influenced by their shape.
Prior work has explored morphologies ranging from tube-shaped designs~\cite{iizuka2024low} and seed-inspired designs~\cite{iyer_wind_2022, wiesemuller_transient_2022, kim_three-dimensional_2021, pounds_samara_2015} to glider-type sensors~\cite{vikrant2018, girardi_biodegradable_2024}.

Actuated mechanisms have also been explored to change shape mid-air, enabling flyers to switch between different descent modes~\cite{Hla_samara_actuator_2022, johnson_solar-powered_2023}.
While these approaches demonstrate partially controllable deployment, they rely on integrated actuators, energy for deformation, and three-dimensional structures.
Such complexity increases fabrication cost, limits stackability, and constrains large-scale deployment.
Moreover, achieving reliable folding into multiple programmed positions is particularly challenging with actuator-based systems, as each configuration requires precise control and additional energy.
A passive mechanism that transforms from a flat, stackable form into an aerodynamic glider—without mid-flight actuation or external power—is therefore highly desirable.

Methods for creating three-dimensional structures from two-dimensional sheets are valued for their ease of fabrication and transport~\cite{narumi_inkjet_2023}.
Approaches include manual folding of printed circuits~\cite{olberding_foldio_2015, yamaoka_foldtronics_2019, dauden_roquet_3d_2016}, PET film~\cite{liu_electropaper_2021}, and rigid boards~\cite{freire_pcbend_2023}, as well as self-folding via heat-driven laminates~\cite{bachmann_self-folding_2022, yan_fibercuit_2022, felton_method_2014, felton_self-folding_2013} or shape-memory polymers~\cite{wang_electrothermally_2024, wang_morphingcircuit_2020, hong_thermoformed_2021}.
Yet no method simultaneously integrates rigid PCBs, provides repeatable elastic transformation, and allows simple 2D fabrication.
\begin{figure}[t]
  \centering
  \includegraphics[width=\linewidth]{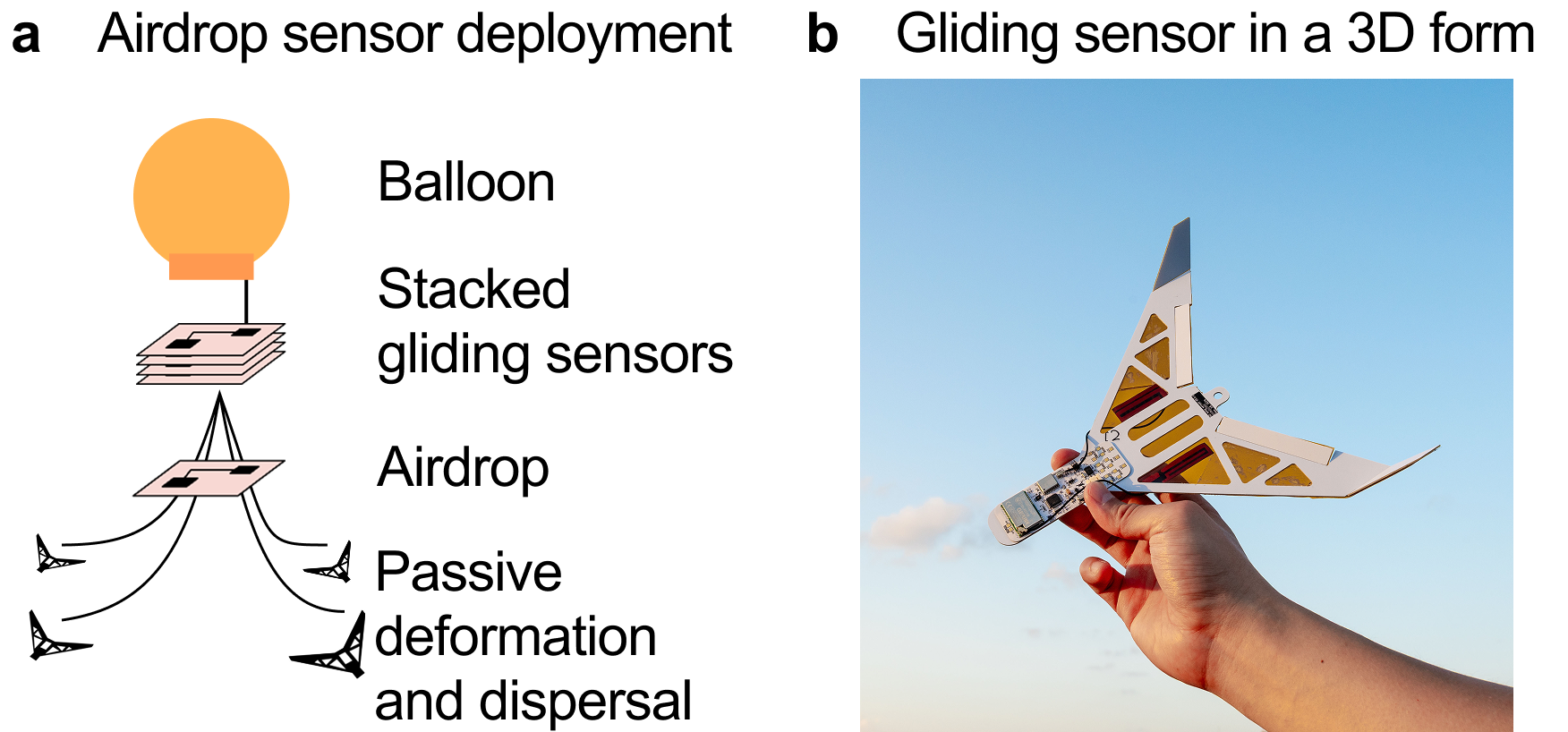}
  \caption{
  Elastic-folding sensor concept and transformation process.
  (a)~The sensor stacked in planar form before deployment.
  (b)~The sensor passively transforms into a 3D glider after dispersion.}
  \label{fig:1}
\end{figure}
\begin{figure*}[tbp]
  \centering
  \includegraphics[width=\linewidth]{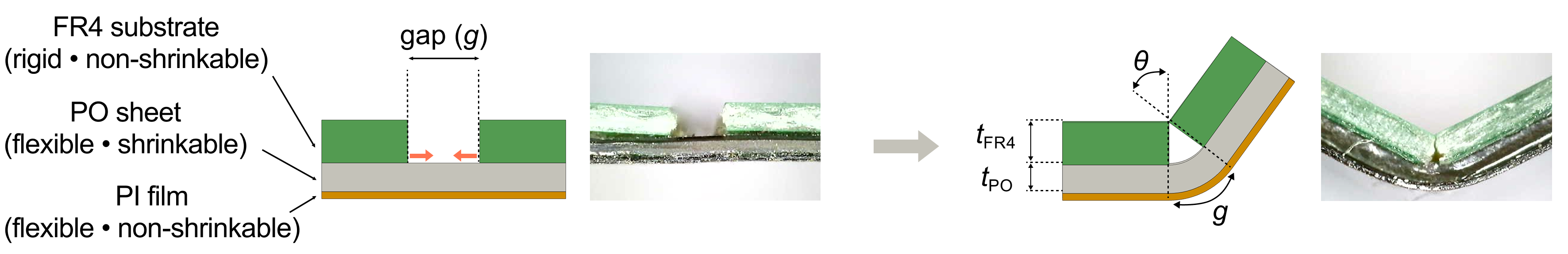}
  \caption{
  Cross-sectional view of the three-layered elastic hinge before and after heating.
  A heat-shrinkable PO sheet is laminated with a gapped FR4 substrate and a polyimide film.
  Upon heating, the exposed PO sheet shrinks, causing the hinge to fold.
  }
  \label{fig:2}
\end{figure*}
This work proposes and evaluates a passive transformation mechanism for air-deployed sensors, motivated by future deployment from radiosonde balloons for weather forecasting.
In this envisioned application, lightweight sensors are released from radiosondes over the ocean, collect atmospheric data during descent, and transmit it via LPWA communication~\cite{iizuka2024low}.
The proposed mechanism enables each sensor to transform from a planar, stackable form into a three-dimensional glider morphology (Fig.~\ref{fig:1}).

The key idea of this work is a passive elastic-hinge mechanism fabricated by laminating a rigid circuit board with a heat-shrink polymer and a flexible sheet, followed by a single heating step.
This approach combines three properties critical for scalable aerial sensing: (1) simple fabrication compatible with electronic integration, (2) elastic recovery from a flat, stackable state to a designed 3D shape, and (3) fully passive transformation without active actuation after release.

We first present the principle and fabrication method for the proposed self-folding hinge, along with a geometric model for predicting fold angles.
We then analyze hinge elasticity, enabled by thicker polymer layers compared to prior studies.
Finally, we implement the mechanism in a glider-shaped wireless sensor prototype, demonstrate reliable data collection during dispersion, and evaluate performance for radiosonde-based airdrop applications through simulation.

The contributions of this work are:
\begin{itemize}
\item Proposing an elastic-hinge mechanism, fabricated by laminating and heating sheet materials with rigid PCBs, with a model for predicting fold angles.
\item An analysis of hinge elasticity with experimental evaluation under repeated deformation and sustained compression.
\item An implementation of the mechanism on a glider-shaped sensor and evaluation of its flight performance for radiosonde-based airdrop sensing.
\end{itemize}

\section{Passive Elastic-Folding Mechanism}
The hinge mechanism for our deployable sensor must fulfill two transformation requirements. First, it requires an initial, one-time heat-triggered self-folding.
This allows us to produce the 3D structured glider efficiently and precisely using 2D fabrication techniques, avoiding the challenges of assembling small, complex shapes. Second, for deployment, it must function as a passive elastic hinge, capable of being stored flat and repeatedly springing back to its functional 3D shape upon release, without requiring power or active components such as actuators. 

This section details (\ref{subsection:principle}) the principle of self-folding and passive elastic folding, (\ref{subsection:mechanism}) the fabrication method of the proposed mechanism, and (\ref{subsection:angle_design_model}) the design model for achieving a desired folding angle.

\subsection{Principle}
\label{subsection:principle}
Self-folding is achieved by heating a three-layer laminate consisting of a rigid FR4 substrate patterned with a gap, a heat-shrinkable and flexible polyolefin (PO) sheet, and a non-shrinkable flexible polyimide sheet, as shown in Fig.~\ref{fig:2}. When heated, only the exposed region of the PO sheet contracts, while the bonded areas remain stable. This shrinkage difference bends the PO and polyimide layers until the rigid edges of the FR4 substrate make contact.

After this initial folding process, the resulting 3D structure functions as an elastic hinge. This is enabled by using a thick elastic PO sheet as the heat-shrinkable layer. Previous self-folding methods~\cite{narumi_inkjet_2023, felton_method_2014} have commonly relied on polystyrene (PS) or thin-film PO. However, these materials lack the properties required for a passive elastic-folding hinge. PS is brittle at or below room temperature and fractures under large deformation, limiting it to one-time plastic folding. Conventional thin PO films ($< \SI{50}{\micro\meter}$) are too compliant to provide sufficient hinge stiffness. To address these limitations, we repurposed a PO heat-shrink tube, cut open into a flat sheet with a thickness of \SI{350}{\micro\meter} (Table~\ref{tab:1}). This material provides both stiffness and elasticity, enabling repeated elastic folding.

\subsection{Fabrication Process}
\label{subsection:mechanism}
\begin{figure*}[t]
  \centering
   \includegraphics[width=\linewidth]{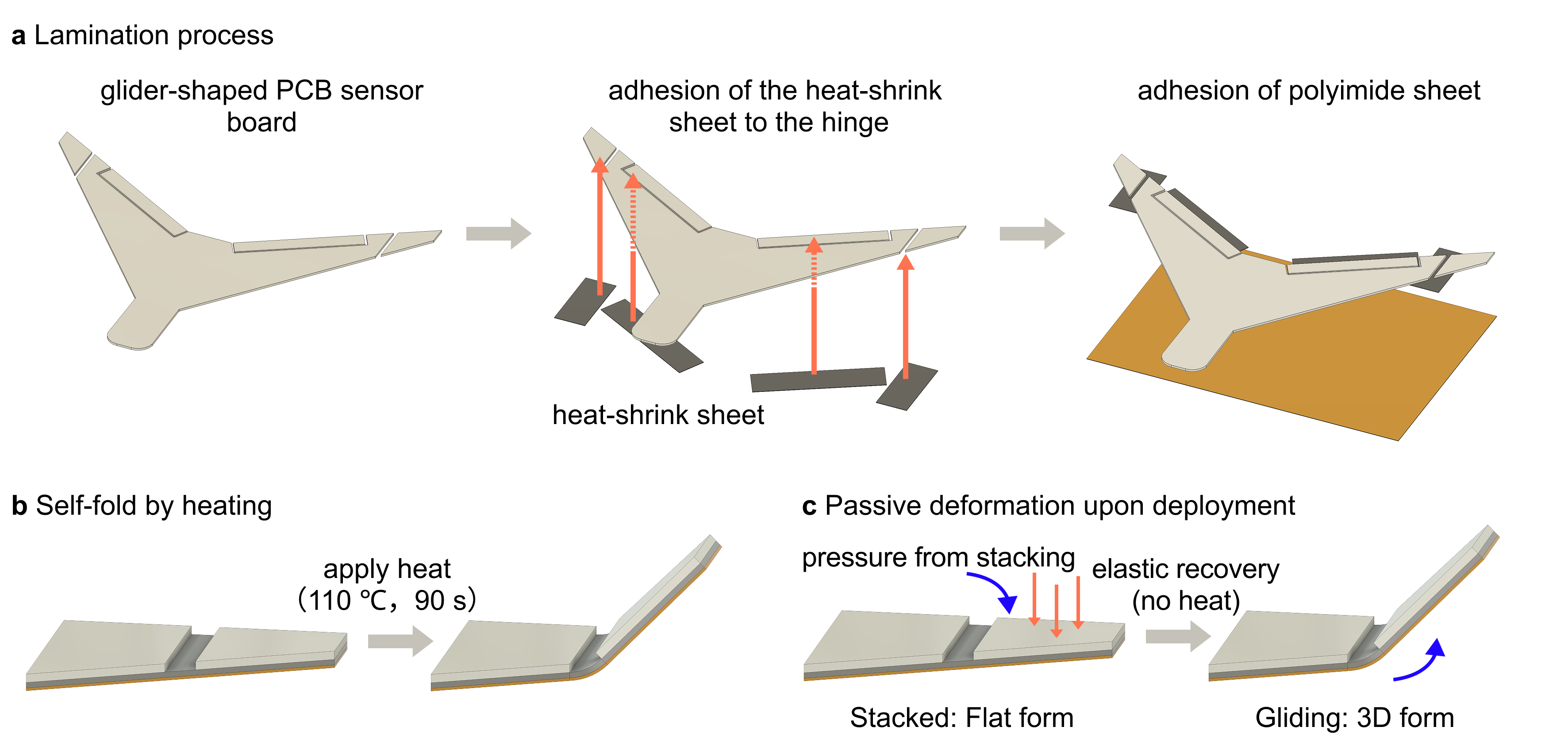}
  \caption{
  Fabrication process and passive transformation of the elastic-folding sensor using the proposed hinge mechanism.
  (a) The fabrication process begins with a rigid PCB patterned with a predesigned gap at the target hinge location. A heat-shrink polyolefin sheet and a polyimide sheet are laminated to the substrate and then heated in a reflow oven.
  (b) During heating, the exposed portion of the polyolefin sheet contracts. The strain mismatch between this layer and the polyimide layer generates a bending moment that folds the hinge to an angle determined by the PCB gap geometry. This heat-driven self-folding process occurs during fabrication, prior to stacking and loading onto the balloon.
  (c) After fabrication, the hinge can be flattened under pressure for stacking. Upon release during deployment, its elasticity enables passive recovery into the designed 3D glider configuration.
  }
  \label{fig:3}
\end{figure*}
The fabrication is completed in three steps as shown in Fig.~\ref{fig:3}: (1) designing the PCB gap corresponding to the required dihedral angle, (2) laminating the three-layer structure, and (3) deforming the structure into its 3D shape by applying heat.

First, based on the hinge design model we analyze in the next section, we design the circuit board gap to achieve the required fold angle.
Second, the heat-shrink PO sheet, polyimide film, and FR4 substrate are laminated together. As shown in Fig.~\ref{fig:2}, the FR4 substrate is patterned with a gap ($g$).
Finally, the entire sheet is heated in a reflow oven. The exposed portion of the PO sheet contracts, and a strain mismatch between this layer and the polyimide layer generates a bending moment that folds the hinge. As we will show, this allows the final angle to be reliably controlled by the gap geometry ($g$). 
After this initial fabrication, the resulting folded hinge can be flattened by applying pressure. When the pressure is released, the structure returns to its original folded angle due to its elastic recovery. This elastic characteristic allows the structure to function as a passive deformation mechanism (Fig.~\ref{fig:3}).

\subsection{Fold Angle Model}
\label{subsection:angle_design_model}
We constructed a geometric model to predict the final folding angle from the initial board gap $g$. When the laminated structure is heated, the differential shrinkage between the upper surface of the heat-shrink material and its lower surface fixed to the polyimide sheet causes the hinge to bend, resulting in self-folding deformation. By assuming that sufficient heating causes the boards to shrink until they make contact, as depicted in Fig.~\ref{fig:2}, the radius of curvature $R$ of the hinge can be expressed as:
\begin{equation}
    \label{eq:R=t+t}
    R = t_{\mathrm{FR4}} + t_{\mathrm{PO}}
\end{equation}
Here, $t_{\mathrm{FR4}}$ is the thickness of the FR4 substrate and $t_{\mathrm{PO}}$ is the thickness of the polyolefin heat-shrink material. The relationship between the radius of curvature $R$, the board gap $g$, and the dihedral angle  (in radians) is given by $g=R\theta$. Therefore, the final folding angle can be expressed as follows:
\begin{equation}
    \theta = \dfrac{g}{t_{\rm FR4} + t_{\rm PO}}
    \label{eq:theta=}
\end{equation}
The value of \(t_{\mathrm{PO}}\) depends on the shrinkage rate under specific heating conditions and was measured after applying a consistent heating process.

\begin{figure}[tbp]
  \centering
  \includegraphics[width=\linewidth]{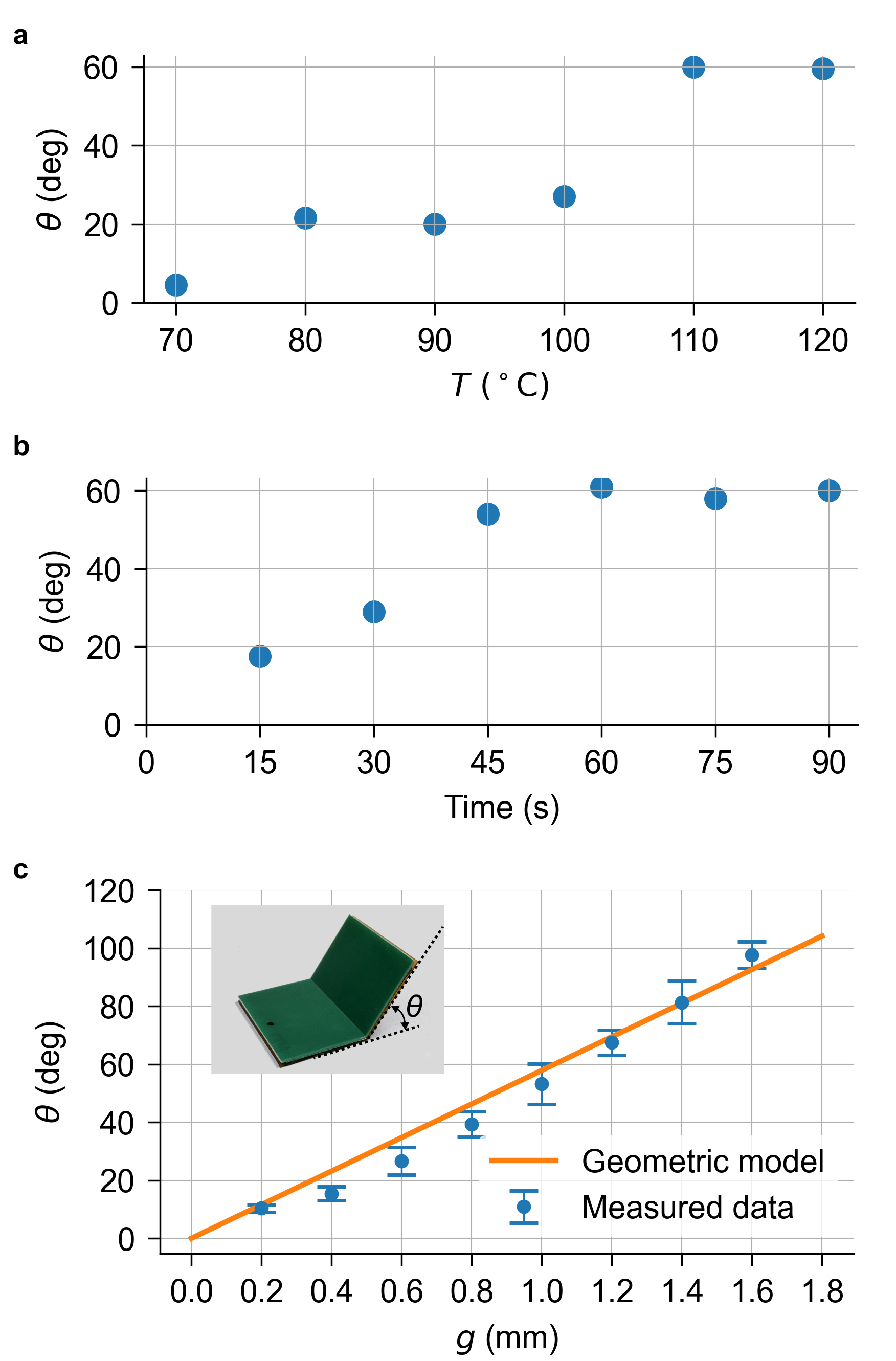}
  \caption{
  Experimental validation of the heating conditions and hinge angle model. 
  (a)~Relationship between heating temperature and fold angle, with heating time fixed at \SI{90}{\second}.
  (b)~Relationship between heating time and fold angle, with heating temperature fixed at \SI{110}{\celsius}.
  (c)~Relationship between the board gap ($g$) and the fold angle. In addition to the measured data, theoretical values calculated using (\ref{eq:theta=}) are shown.
  }
  \label{fig:4}
\end{figure}

\section{Evaluation of Elastic-Folding Hinge}
\begin{figure}[tbp]
  \centering
  \includegraphics[width=\linewidth]{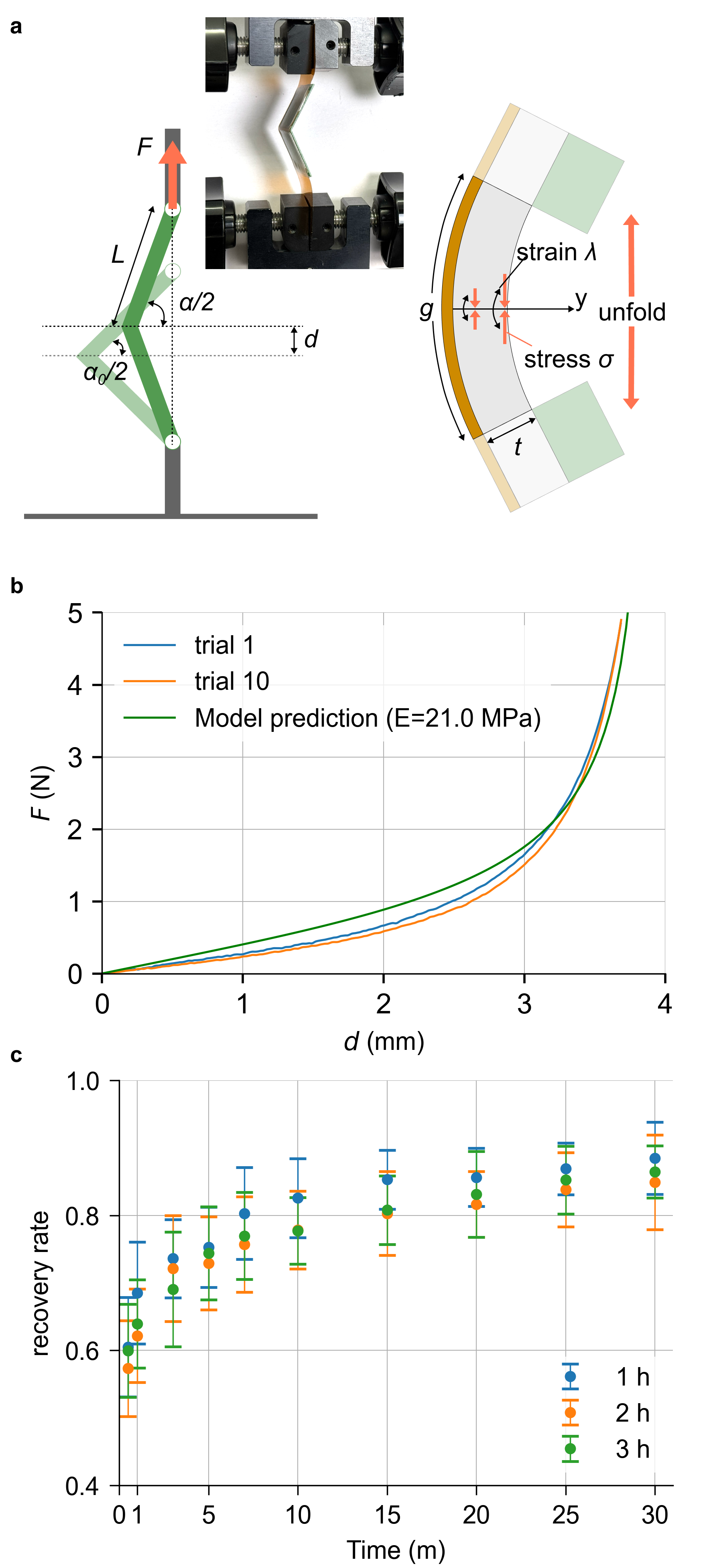}
  \caption{
  Elastic hinge evaluation under cyclic deformation.
  (a)~The tensile test setup and a diagram of the stress in the hinge. 
  A force is applied to unfold the hinge.
  This results in greater extension and higher stress on the inner side of the hinge.
  (b)~Load-displacement curve from the tensile test.
  The graph shows the results for the 1st and 10th cycles, along with the theoretical curve.
  (c)~Relationship between elapsed time (in minutes) after release from compression and the angle recovery rate.
  Samples were held in a planar state under pressure for 1 to 3 hours before release.
  }
  \label{fig:5}
\end{figure}

\begin{table}[tbp]
    \centering
    \caption{Sheet Materials}
    \label{tab:1}
    \begin{tabular}{@{}lll@{}}
    \toprule
    Material & Specification / Part No. & Thickness (\si{\micro\meter}) \\ \midrule
    Polyolefin Sheet  & THT-14.0 N & 350 \\
    Polyimide Sheet   & 200H-A4    & 50 \\
    FR4 Substrate    & 0.6~mm PCB (JLCPCB) & 600 \\ \bottomrule
  \end{tabular}
\end{table}

In this section, we first determine the optimal heating conditions and then validate the hinge model presented in the previous section.

\subsection{Heating Conditions}
We first investigated the heating conditions required to achieve sufficient thermal shrinkage without causing structural failure. Excessively high temperatures risk adhesive delamination, while insufficient temperatures result in incomplete folding. Using the model from the previous section, the gap was designed to produce a 60$^\circ$ angle, and the relationship between heating time, temperature, and the resulting dihedral angle was examined. The relationship between heating temperature and dihedral angle is shown in Fig.~\ref{fig:4}a, and the relationship between heating time and dihedral angle is shown in Fig.~\ref{fig:4}b. The results confirmed that heating at \SI{110}{\celsius} for \SI{90}{\second} or more is sufficient to achieve the designed dihedral angle. It was also observed that heating at \SI{120}{\celsius} or higher tended to cause structural failure due to reduced adhesive strength. From these results, all the following experiments were conducted using a heating condition of \SI{110}{\celsius} for \SI{90}{\second}. This temperature is also significantly lower than that of standard reflow soldering. This ensures that pre-mounted electronic components are not damaged during the self-folding process.

\subsection{Validation of the Fold Angle Model}
To validate the fold angle model presented in Section~\ref{subsection:angle_design_model}, we evaluated the hinge's dihedral angle while varying the PCB gap. The resulting dihedral angles for PCB gaps ranging from \SI{0.2}{\milli\meter} to \SI{1.6}{\milli\meter} in \SI{0.2}{\milli\meter} increments are shown in Fig.~\ref{fig:4}c. Five samples were used for each gap value. The figure also shows the relationship based on the analytical model from (\ref{eq:theta=}), using the measured board thickness \(t_{\mathrm{FR4}}(=\SI{0.55}{\milli\meter})\) and the post-shrinkage thickness of the heat-shrink material \(t_{\mathrm{PO}} (=\SI{0.44}{\milli\meter})\). The predicted values from the model showed strong agreement with the measurement results (coefficient of determination \(R^2 = 0.93\)).  Furthermore, the standard deviation of the absolute error between the measured angles and the model was \ang{4}.

This evaluation confirmed that by setting the gap $g$ using the model developed in (\ref{subsection:angle_design_model}), a hinge structure with a desired dihedral angle can be fabricated reliably and with high precision.

\subsection{Force-Displacement Characteristics}

To characterize the elastic properties of the hinge, we evaluated its force-displacement relationship and its durability against repeated deformation. Using a tensile tester, the fabricated hinge was held at both ends and subjected to cyclic folding tests (Fig. \ref{fig:5}a). During these tests, we measured the force required to change the displacement and examined how the force-displacement curve changed over multiple cycles. Finally, we validated the theoretical hinge model by comparing this experimental curve with the one predicted by the model.

\subsubsection{Derivation of force-displacement model}
We theoretically derive the force-displacement response, \(F(d)\), in a tensile test by modeling the hinge section as a Neo-Hookean elastomer with thickness \(t\) and width \(W\). The Neo-Hookean constant \(G\) is the shear modulus.
For a hinge with substrate length \(L\) and an initial fold angle of \(\theta\) (\(\alpha_{0}\ =\pi-\theta_0\)), the internal angle \(\alpha\) when unfolded by a displacement \(d\) in a tensile tester is given by:
\begin{equation}
\begin{aligned}
d &= 2L(\sin\frac{\alpha}{2} - \sin\frac{\alpha_{_0}}{2})  \\
\alpha &= 2
\arcsin( \frac{d}{2L} + \sin\frac{\alpha_{_0}}{2})
\end{aligned}
\label{eq:alpha}
\end{equation}
The bending moment \(M\) acting on the hinge is expressed as:
\begin{equation}
M = F L \cos\frac{\alpha}{2}
\label{eq:M=}
\end{equation}
Here, \(F\) is the tensile force. Assuming the inner side of the hinge is the tensile side and that there is no extension on the polyimide side, the gap \(g\) on the polyimide side is fixed. We assume it deforms into a circular arc with a pure bending curvature \(\varphi\). The curvature is then \(\varphi = \theta/g\). Let the thickness in the planar state be \(t_\mathrm{flat}\), with the coordinate in the thickness direction being \(Y \in [0,t_\mathrm{flat}]\).

From the incompressibility of the material, the following relationships hold:
\begin{equation*}
\begin{aligned}
Y\cdot g &= \frac{1}{2}(R^2 - (R-y)^2)\theta \\
y &= \frac{1 - \sqrt{1 - 2Y\varphi }}{\varphi}
\end{aligned}
\end{equation*}
At this time, the stretch ratio \(\lambda\) relative to the planar state and the stress \(\sigma(y)\) at position \(y\) are expressed by the Neo-Hookean model as:
\begin{equation*}
\begin{aligned}
\lambda &= 1 - y\varphi  =  \sqrt{1 - 2Y \varphi}\\
\sigma(y) &= G( \lambda - \lambda^{-2} )
\end{aligned}
\end{equation*}
When the fold angle is reduced from the initial angle \(\theta_0\) to \(\theta_1\) by tension, the effective stretch is \(\lambda_{eff} =  \frac{\lambda_{1}}{\lambda_{0}}\), and the bending moment \(M\) is:
\begin{equation}
\begin{aligned}
   M &= W \int_0^{t_{1}} \sigma(y)\, y \, dy \\
   &= GW\int_0^{t_\mathrm{flat}} (\lambda_{eff} - {\lambda_{eff}}^{-2})(\frac{1 - \lambda_{1}}{\varphi_{1}})(\frac{1}{\lambda_{1}})dY 
\end{aligned}
\label{eq:Moment}
\end{equation}
\(M\) is obtained by calculating this integral. Therefore, the tensile force \(F(d)\) is obtained by simultaneously solving (\ref{eq:alpha}), (\ref{eq:M=}) and (\ref{eq:Moment}):
\begin{equation*}
F(d) = \frac{M(d)}{L \cos\frac{\alpha(d)}{2}}
\end{equation*}

\subsubsection{Tensile test}
In the tensile test, a hinge sample with a gap (\(g\)) of \SI{1}{\milli\meter} and a fold angle of \ang{51} (initial internal angle \(\alpha_0 = \ang{129} \)) was subjected to 10 repeated deformation cycles, which exceeds the expected number of folding events—including post-fabrication testing and a one-time planar-to-3D deformation upon mid-air deployment. The results are shown in Fig.~\ref{fig:5}b. The theoretical curve from the hinge model showed a high degree of agreement with the measured results. Specifically, when the Young's modulus (\(E\)) in the model was used as a fitting parameter and applied to the load-displacement data of the first trial using the least-squares method, we obtained a value of \(E = \)\SI{21}{\mega\pascal}. The coefficient of determination for this fit was \(R^2 = 0.97\). This suggests that the model can be used to design the hinge's moment.

\subsection{Elastic Recovery After Long-term Stacking}
Long-period stacking may be necessary when deploying sensors from remote, high-altitude locations. Therefore, we evaluated the hinge's ability to recover to its 3D state after being held in a planar form. Specifically, we used clamps to hold elastic hinge samples flat to assess the angle recovery characteristics over time after release.

Elastic hinge samples with a gap (\(g\)) of \SI{1}{\milli\meter} were held flat under pressure for 1, 2, and 3 hours at room temperature, and the change in the fold angle (\(\theta\)) upon recovery was measured. We tested five samples for each stacking duration. A typical radiosonde balloon maintains a steady ascent rate and reaches an altitude of \SI{10}{\kilo\meter} within \SI{1}{\hour}~\cite{iizuka_balloon_2025}.
We set the \SI{3}{\hour} stacking duration for evaluation, considering these real-world factors. The results are shown in Fig.~\ref{fig:5}c. 
Notably, 30 minutes after release, the angle recovery rate was consistently around 85\% regardless of the duration of flat holding.

The hinge's recovery process comprises two phases: a large, instantaneous elastic recovery immediately after release, followed by a slower, time-dependent viscoelastic recovery. For our application, the initial recovery within 30 minutes is the most critical factor, as this corresponds to the glider's typical descent time.

Our results show that even after being held flat for up to three hours—a sufficient duration for a radiosonde balloon to reach its deployment point over the ocean—the recovery rate stabilizes at a predictable rate. This predictability implies that the desired fold angle can be achieved by accounting for the recovery rate. Furthermore, we hypothesize that at high altitudes, where temperatures are often below freezing, the time-dependent recovery would be suppressed, making the initial elastic recovery even more dominant.

\section{Sensor Prototype and Flight Evaluations}
This section presents the design and prototyping of a gliding sensor airframe integrated with sensor circuit boards. We conducted outdoor flight tests and simulated the trajectory of the deployed system from a high altitude. 

\subsection{Prototype Glider Sensor}
\label{subsection:prototype}
\begin{figure}[tbp]
  \centering
    \includegraphics[width=\linewidth]{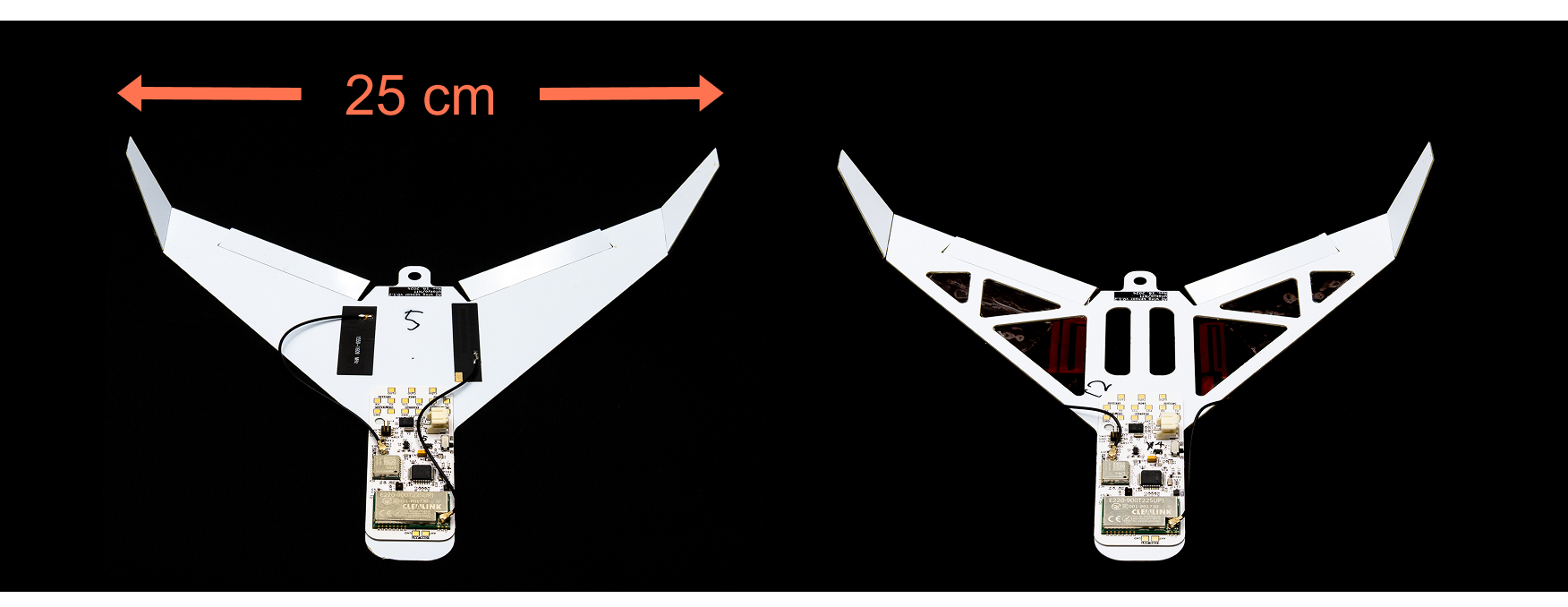}
    \caption{
    Glider sensor prototypes for environmental sensing.
    Two versions were developed: a standard design (left) and a cutout design (right), both with \SI{25}{\centi\meter} wingspans and integrated PCBs with sensors and LoRa communication.
    }
  \label{fig:6}
\end{figure}

We developed two prototype airframes with a \SI{25}{\centi\meter} wingspan for environmental sensing: a standard design and a cutout design (Fig.~\ref{fig:6}). Both versions consist of a printed circuit board (PCB) with temperature, humidity, GPS, and IMU sensors, a battery, and a LoRa communication module, thus achieving a full integration of the sensor electronics with the glider airframe.

\subsection{Flight Tests in an Outdoor Robotics Test Field}
\label{subsection:rtf}
We tested the designed airframe's capability as an environmental glider sensor by conducting launch tests in an outdoor environment.

\begin{figure}[tbp]
  \centering
  \includegraphics[width=\linewidth]{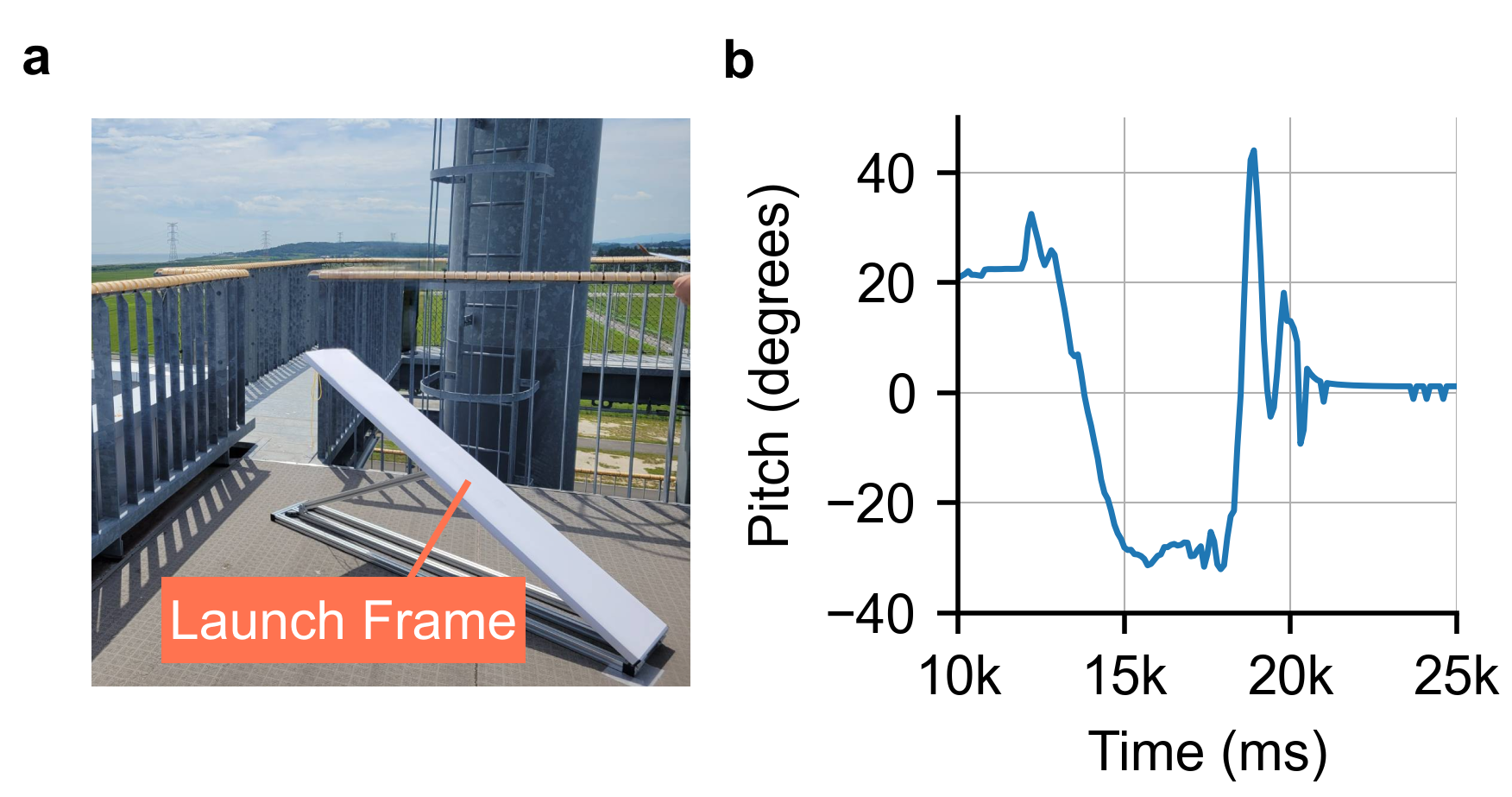}
  \caption{
  Outdoor dispersion test of the glider sensor at the Fukushima Robot Test Field.
  The airframe integrates a sensor board with IMU, GPS, environmental sensors (temperature and humidity), and a LoRa communication module. 
  (a)~Launch setup from a \SI{30}{\meter} platform using a custom-built frame and rubber band. 
  (b)~Onboard IMU pitch-angle data capturing the passive pull-up maneuver after launch.
  }
  \label{fig:7}
\end{figure}

The tests were conducted at the Fukushima Robot Test Field, where the airframe was launched from a height of 30 meters to observe its flight characteristics. To ensure consistent launches, we used a custom-built aluminum frame to launch the prototype with a rubber band (Fig.~\ref{fig:7}a). 

During the flight, we successfully received data packets from all onboard sensors via the LoRa communication link, confirming the viability of the integrated system. As shown in Fig.~\ref{fig:7}b, analysis of the IMU data confirmed that the airframe passively performed a pull-up maneuver after launching from a height of 30 meters. This suggests that the airframe will also be able to transition to a stable glide upon deployment from its target altitude of \SI{10}{\kilo\meter}, while acquiring and transmitting sensor data.

\begin{figure}[tbp]
  \centering  \includegraphics[width=\linewidth]{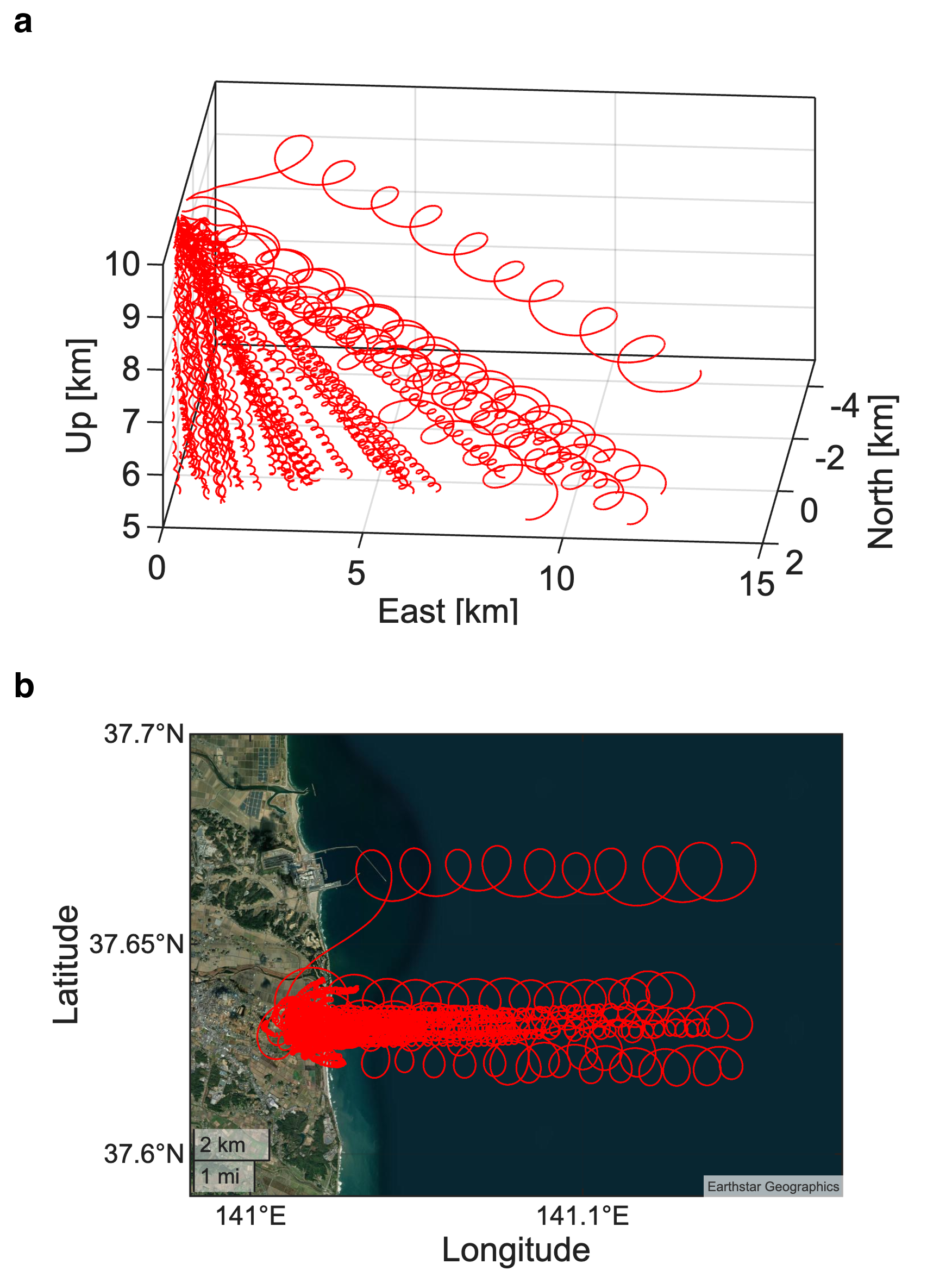}
  \caption{
  Dispersion simulation of glider airframes deployed from high altitude.
  (a)~3D trajectories of 50 glider airframes released from an altitude of \SI{10}{\kilo\meter}, showing descent in turning patterns while drifting with the wind. 
  (b)~Top-down view of the simulated trajectories over Mito, Japan, showing the descent from a \SI{10}{\kilo\meter} deployment altitude to the simulation's end altitude of \SI{5}{\kilo\meter}.
  The wind environment combines a mean wind profile from MATLAB's Horizontal Wind Model (\texttt{atmoshwm}) with turbulence generated by the Dryden gust model.}
  \label{fig:8}
\end{figure}

\subsection{Dispersion Simulation}
To evaluate how the proposed hinge fabrication method affects flight performance from high-altitude deployment, we simulated the trajectory of the gliding sensor when deployed from a high altitude. The simulation's wind environment replicated the steady winds and gusts over Mito, Japan—the planned site for future field tests—based on local meteorological data.

Slight variations in the hinge fabrication process can lead to asymmetrical variations in the fold angles of the left and right elevons. To investigate the effect of these variations on the flight path, we conducted a Monte Carlo simulation, deploying numerous airframes with slightly different elevon angles from an altitude of \SI{10}{\kilo\meter}.

The results confirmed that many of the airframes descended in a turning pattern due to the difference in their left and right angles (Fig.~\ref{fig:8}). The turning radius for each airframe varied depending on its specific angle deviation, causing the final landing positions to be dispersed over a wide area with a diameter of \SI{10}{\kilo\meter}.

This outcome suggests that the angle control afforded by our hinge design can be applied to achieve wide-area sensor dispersal. Unlike a conventional parachute with a relatively constant hang time, our method produces large variations in the turning radius—and consequently, the hang time and wind-drift distance—for each airframe. By leveraging this characteristic, effective dispersal over a wide area can be achieved from a single deployment point.

\section{Discussion}
Prior efforts to realize functional three-dimensional electronic structures have largely used 3D-printed circuits~\cite{yan_fibercuit_2022, wang_electrothermally_2024, wang_morphingcircuit_2020, hong_thermoformed_2021} or heat-shrink laminates~\cite{felton_self-folding_2013}. Integrating electronics into 3D-printed bodies typically requires complex manual assembly and delivers lower performance than commercial rigid PCB workflows. By contrast, most heat-shrink approaches depend on one-time plastic deformation, limiting elasticity and preventing passive refolding at deployment.

Our approach addresses these gaps by combining three properties essential for deployable sensing: direct integration of commercial rigid circuit boards, repeatable elastic deployment, and straightforward 2D manufacturing. The elastic-hinge mechanism reliably converts stackable, planar devices into gliders for sensing and wide-area dispersion. Although recovery to the designed folding state is not instantaneous and fold-angle errors remain, aerial simulations indicate that these deviations preserve long-range travel and can even diversify sampled trajectories.

\subsection{Future work}
Future efforts will prioritize a rigorous aerodynamic characterization of the platform in both steady-state and transient regimes. While the present work establishes the hinge mechanics and provides initial flight results, a comprehensive study is required to map performance metrics against the continuous geometric evolution seen during deployment. Because the elastic folding recovery can take up to 30 seconds to reach 70\% of the nominal geometry, developing designs that explicitly account for this shape-morphing phase is a critical next step. Integrating these aerodynamic insights with high-throughput manufacturing will allow for the optimization of glider architectures for predictable, large-scale environmental monitoring.

Parallel to these aerodynamic studies, we aim to investigate hinge performance across the broader envelope of high-altitude deployment conditions, specifically focusing on hypobaric and cryogenic environments. While sub-freezing temperatures may suppress viscoelastic relaxation and potentially enhance initial elastic recovery, the coupled effects of thermal stiffening and low-density air on deployment stability present an important area for optimization. Characterizing these environmental interactions will be essential for ensuring consistent mechanical performance and structural reliability across diverse atmospheric profiles.

\section{Conclusions}
We presented a passive elastic-hinge mechanism that transforms sensors from a planar, stackable form into a three-dimensional gliding morphology. The hinge is fabricated by laminating a rigid circuit board, heat-shrink polymer, and flexible sheet, followed by a single heating step. A geometric model predicts fold angles from laminate parameters, and experiments confirm accuracy, repeatability, and elastic recovery. Field tests demonstrated integrated prototypes that transmitted IMU, GPS, and environmental data via LoRa during dispersion, validating sensing and communication under realistic conditions. Collectively, these results establish a simple, scalable, and PCB-compatible folding mechanism as an enabling technology for wide-area environmental monitoring.

\addtolength{\textheight}{-1cm}   





\bibliographystyle{IEEEtran}
\bibliography{ICRA2026}

\end{document}